Marmik Shrestha, Omar Hisham Alsadoon, Abeer Alsadoon*, Thair Al-Dala'in, Tarik A. Rashid, P.W.C. Prasad, Ahmad Alrubaie (2022). A novel solution of deep learning for enhanced support vector machine for predicting the onset of type 2 diabetes, Multimedia Tools and Applications.
DOI : https://doi.org/10.1007/s11042-022-13582-9

# A Novel Solution of Deep Learning for Enhanced Support Vector Machine for Predicting the Onset of Type 2 Diabetes

Marmik Shrestha[1], Omar Hisham Alsadoon[2], Abeer Alsadoon[1,3,4,5*], Thair Al-Dala'in[1,3,4,5], Tarik A. Rashid[6], P.W.C. Prasad[1,3,4], Ahmad Alrubaie[7]

[1] School of Computing Mathematics and Engineering, Charles Sturt University (CSU), Australia
[2] Department of Islamic Sciences, Al Iraqia University, Baghdad, Iraq
[3] School of Computer Data and Mathematical Sciences, Western Sydney University (WSU), Sydney, Australia
[4] Kent Institute Australia, Sydney, Australia
[5] Asia Pacific International College (APIC), Sydney, Australia
[6] Computer Science and Engineering Department, University of Kurdistan Hewler, Erbil, KR, Iraq.
[7] Faculty of Medicine, University of New South Wales, Sydney, Australia

Abeer Alsadoon[1*]
* Corresponding author. A/Prof (Dr) Abeer Alsadoon, [1]School of Computing and Mathematics, Charles Sturt University, Sydney Campus, Australia. Email: alsadoon.abeer@gmail.com , Phone +61 2 9291 9387

## Abstract

Type 2 Diabetes is one of the most major and fatal diseases known to human beings, where thousands of people are subjected to the onset of Type 2 Diabetes every year. However, the diagnosis and prevention of Type 2 Diabetes are relatively costly in today's scenario; hence, the use of machine learning and deep learning techniques is gaining momentum for predicting the onset of Type 2 Diabetes. This research aims to increase the accuracy and Area Under the Curve (AUC) metric while improving the processing time for predicting the onset of Type 2 Diabetes. The proposed system consists of a deep learning technique that uses the Support Vector Machine (SVM) algorithm along with the Radial Base Function (RBF) along with the Long Short-term Memory Layer (LSTM) for prediction of onset of Type 2 Diabetes. The proposed solution provides an average accuracy of 86.31% and an average AUC value of 0.8270 or 82.70%, with an improvement of 3.8 milliseconds in the processing. Radial Base Function (RBF) kernel and the LSTM layer enhance the prediction accuracy and AUC metric from the current industry standard, making it more feasible for practical use without compromising the processing time.

## 1. Introduction

Diabetes is a chronic disease that causes serious health implications and affects millions of people of all age groups worldwide [1]. Diabetes has been clinically classified into two types, namely, Type 1 Diabetes and Type 2 Diabetes. Type 1 Diabetes happens when the human body cannot make any insulin, while Type 2 Diabetes happens when the body cannot produce enough insulin [2]. The proposed system will deal with the prediction of the onset of Type 2 Diabetes, and the prediction of the onset of Type 1 Diabetes is out of scope for the proposed solution. Type 2 Diabetes is a widespread metabolic disease, with thousands of people suffering from the disease worldwide [2]. The number of Type 2 Diabetes patients is increasing at an alarming rate every year. During the early days of medical research acceleration, the Glycated Hemoglobin (A1C) tests, fasting plasma glucose tests, oral glucose tolerance tests were performed to calculate the blood sugar level for a specific time frame [3].

The prediction of onset of Type 2 Diabetes has been facilitated by using the deep learning technique, which has helped increase prediction accuracy compared to other machine learning approaches [4]. Deep learning can speed up and enhance the accuracy for predicting the onset of Type 2 Diabetes and reducing the resources and tests that need to be conducted whilst using the traditional method [5]. Deep learning can repurpose the health record that has been used for other purposes and use the feature sets of the subjects in the health record for predicting the onset of disease on a large volume of people with minimal effort [6].

This research aims to increase the accuracy and Area Under the Curve (AUC) metric and improve the processing time for predicting the onset of Type 2 Diabetes. The proposed system consists of a deep learning technique that uses the Support Vector Machine (SVM) algorithm along with the Radial Base Function (RBF) along with the Long Short-term Memory Layer (LSTM) for prediction of onset of Type 2 Diabetes. Likewise, the process uses a Multi-Layer Perceptron (MLP) to improve the (AUC metric) prediction. The industry standard diabetes prediction using machine learning techniques do not have a satisfactory prediction accuracy level [7]. Hence, the proposed system uses the RBF kernel coupled with the SVM algorithm to yield better prediction accuracy than the current industry standard machine learning approaches while keeping the processing time for each subject in check and improving the processing time.

Many different machine learning approaches are currently being implemented to predict the onset of Type 2 Diabetes, such as deep learning, Artificial Neural Network (ANN). These approaches have been tested to predict the onset of Type 2 Diabetes, and deep learning has proven to be the most effective and efficient method for the prediction, providing more accurate results across the board in different scenarios. It enables healthcare professionals to predict diabetes in a given subject using intricate networks with high accuracy [8].

The rest of the paper is organized as follows. Section 2 presents a literature review. Section 3 describes the proposed system. Section 4 presents the results. Section 5 presents the discussion. Section 6 presents the conclusion and provides recommendations for future work.

## 2. Literature Review

The process of deep learning usually involves feature extraction, feature analysis and pre-processing before the deep learning technique can interpret the feature sets and provide the analysis on the extracted feature sets, which in turn helps predict the onset of Type 2 Diabetes [9]. The deep learning technique requires a dataset to process and extract the desired features [10]. Bernardini et al. [1] and Nguyen et al. [8] make use of the electronic health record (EHR) which are publicly available government recorded datasets. Likewise, Zou et al. [4], Samant and Agarwal [9] and Raschka et al. [12] use the Pima Indians Datasets as a basis for data pre-processing and feature extraction for prediction of the onset of Type 2 Diabetes. The datasets are then subjected to feature extraction, where the features that are recorded in the datasets are extracted. Therefore, the features can be prepared for selection, and the features that provide the optimum result for the prediction will be analyzed and selected [13]. Kannadasan et al. [14] proposed a deep learning framework to classify the Pima Indians diabetes dataset. Features extracted from the dataset using stacked autoencoders and the dataset classified using softmax layer. The achieved classification accuracy of 86.26%.

Once the datasets are pre-processed and the feature extraction has been performed, the feature sets obtained from the extraction is subjected to feature selection. Different approaches for feature extraction were used with many variations in the feature selection. Body Mass Index (BMI) has been selected as an important feature by [4], [11], [16], [12], [15] and [17], who have extracted the feature from the feature list so that it can be used to analyze the subject and predict the onset of Type 2 Diabetes. Likewise, prior diabetic history is one of the major features that the authors considered for predicting the onset of Type 2 Diabetes. Prior diabetic history was considered as an important feature for diabetes prediction by [4], [11], [18], [9] and [8], which seemed to be one of the most important factors in the prediction of the onset of Type 2 Diabetes in any given subject. Prior diabetic history makes a subject more susceptible to diabetes again, and most of the authors have acknowledged this. Other authors have identified pregnancy condition to be one of the major contributing factors for the onset of diabetes in any given subject [17]. Pregnant women are a high-risk demographic that are more susceptible to diabetes, making it one of the most important features that need to be considered by the authors for predicting the onset of Type 2 Diabetes [6].

The feature analysis phase has seen the most variation. Within deep learning techniques, various algorithms and deep learning approaches predict the onset of Type 2 Diabetes. Bernardini et al. [1] proposed a Sparse Balanced Support Vector Machine (SB-SVM) for feature analysis and classification. The selected features divided into two different feature sets, training dataset, testing and validation datasets. The training dataset was used to train the algorithm with the selected feature set, and the testing and validation datasets were used to perform the actual testing on the technique which Decision Support System has supported. [16], [21], [9] and [15] have all used different variations of SVM for feature analysis and classification. Xie et al. [15] have used Principal Component Analysis (PCA) as the main technique for feature analysis and classification, and they compared the performance of PCA against industry-standard deep learning techniques like Naïve Bayes, Decision Tree and Random Forest. Also, they compared the outcome of the proposed technique against industry-standard techniques. Once the feature has been analyzed and classified, the final model's performance needs to be evaluated in terms of accuracy and processing time. Specificity, Accuracy, Sensitivity and AUC are the metrics that help to weigh the performance of any given technique. Most of the studies focused on a single metric that has seen improvement against the current industry standard. These metrics are calculated by collecting false positive, false negative, true positive and true negative values from the proposed solution's results [1].

## 3. State of the Art

Bernardini et al. [1] proposed a SB-SVM for predicting the onset of Type 2 Diabetes which uses a publicly available FIMMG electronic health record as a dataset for predicting the onset of Type 2 Diabetes on the subjects. It provides acceptable accuracy and AUC metrics that are above the industry standard. Hence, the state of the art [1] was chosen as a basis for our solution, which can be further enhanced to increase the accuracy and AUC metric for prediction of onset of Type 2 Diabetes.

The good features of the current state of the art [1] are highlighted with blue color in Figure 1, and the limitations are highlighted in red color. The use of the SB-SVM algorithm has resulted in the increment in the accuracy with an acceptable processing time of detection of Type 2 Diabetes from a standard medical dataset in comparison to the fellow competitors [22]. The use of the SB-SVM algorithm has resulted in an accuracy of 68.4%. This model provides a promising opportunity for the use of deep learning to predict the onset of Type 2 Diabetes in real scenarios in real life. State of the art consists of three stages, as shown in Figure 1.

In the first stage of the solution, the Federazione Italiana Medici DI Medicina Generale (FIMMG) dataset is obtained from the Metmedica Italia (NMI). The data is pre-processed based on the three subcategories of the data types: Demographic, Monitoring and Clinical, with 1862 feature sets. In the second stage, the pre-processed 1862 feature sets are analyzed, and the necessary feature sets to test the algorithms are isolated from the datasets and the discarded features are kept separately for each of the test cases.

The feature sets are divided into test dataset, training, and validation dataset in the third stage. The test dataset is used to test the algorithm's performance initially, and the rest of the feature set is used to train the algorithm and test the validation of the algorithm in predicting Type 2 Diabetes amongst the provided datasets. The same procedure is followed and reperformed for the remaining two cases, and the accuracy, recall, and AUC are recorded for the subsequent cases.

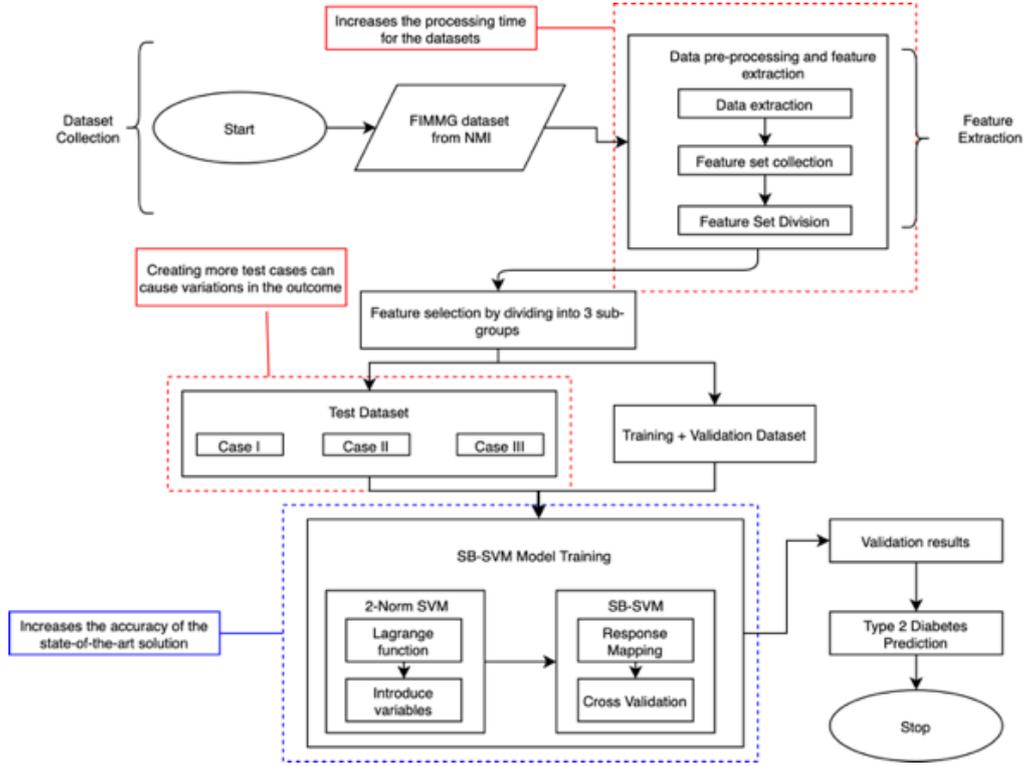

Figure 1. This figure depicts a) the workflow of Type 2 Diabetes Prediction [1]; and b) the good features (depicted by blue colour) and the limitations (depicted by red colour) of the current Type 2 Diabetes prediction using Deep Learning.

The SB-SVM algorithm is implemented in the feature analysis phase in order to deal with the unbalanced setting, which changes the decision threshold directly of the inferred posterior probability while controlling the True positive/ negative rate [1]. Equation 1 shows the activation equation for the state-of-the-art solution.

$$S = \min_{w,w_0,\xi} \frac{1}{2}||w||^2 + C \sum_{i=1}^{M} \xi_i \quad (1)$$

Where,
S = Loss function
$\xi_i$ = slack variable
$w_0$ = bias parameter
$C$ = box constraint and controls the overlapping between the two classes
$\frac{1}{2}||w||^2$ = smallest distance between the decision boundary and any of the training points
$\sum_i m = 1 \ (x_i + y_i)$ = training dataset of m observations and n features

## 4. Proposed System

Bernardini et al. [1] solution was chosen to be as a basis of our proposed solution because of the use of SB-SVM. The SB-SVM algorithm enables the optimization in selecting a feature set and provides better effectiveness and interpretability for the dataset compared to other deep learning methodologies. Bernardini et al. [1] solution perform extremely well in highly unbalanced data settings without employing additional sampling strategies, unlike other deep learning modules. However, the technique's accuracy needs to be improved for better prediction of the onset of Type 2 Diabetes and the AUC metric obtained from the model. Therefore, proposed system aims to increase the accuracy and AUC metric and improve the processing time for predicting the onset of Type 2 Diabetes. The proposed system consists of a deep learning technique that uses the SVM algorithm, the RBF, and the LSTM to predict the onset of Type 2 Diabetes. Likewise, the process uses an MLP to improve the AUC metric prediction, as shown in Figure 2. The RBF kernel is introduced to the SB-SVM to streamline the cross-validation accuracy of each of the subjects [10]. This is an entirely new feature adapted from Samant and Agarwal [9] solution. The proposed solution helps to reduce cross-validation errors and helps to improve the overall accuracy of the algorithm.

The proposed system consists of three major stages, as shown in Figure 2: feature extraction and feature selection, analysis, and diabetes prediction.

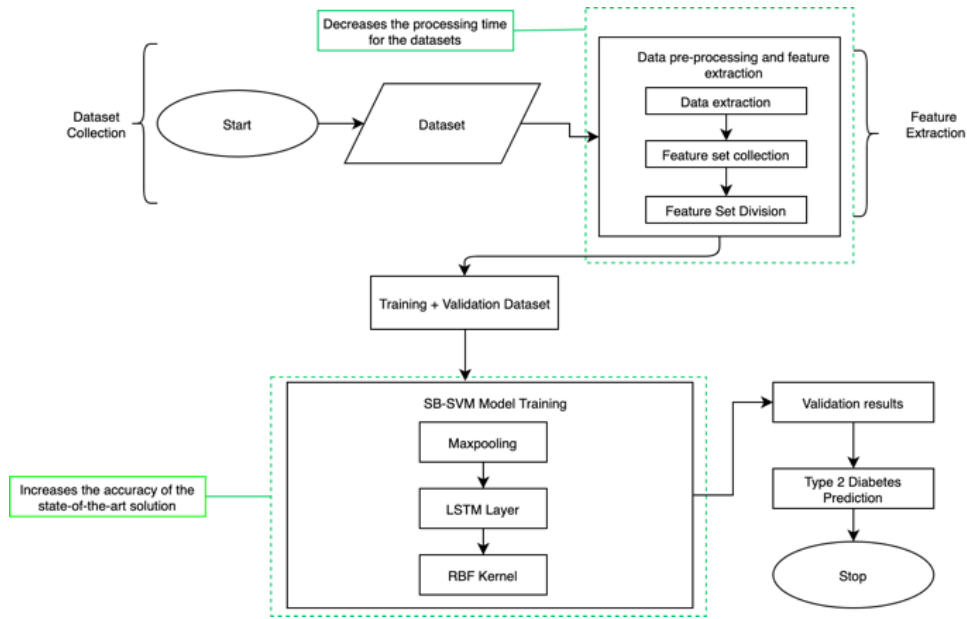

Figure 2 Block diagram of the proposed Deep Learning technique for prediction of onset of Type 2 Diabetes [The green borders refer to the new parts in the proposed system]

**Feature extraction and selection:** During this stage, the data obtained from the dataset is extracted using an extraction technique to identify feature sets included in the provided data sets. The extracted data is divided based on the feature set available in the dataset. The feature sets are divided into different groups based on their property, where the feature set with similar properties are placed in the same group [14]. As shown in Figure 2, after the dataset division, the feature sets used in the prediction of onset of type 2 diabetes are selected for further analysis. The feature sets selected from the provided data set are divided into two groups: Test dataset, training, and Validation dataset. The selected feature set is divided as 30% for the test dataset and 70% for the training and validation dataset. The test dataset is used to initially calculate the module's effectiveness, which can enhance the algorithm's effectiveness when being used on the training and validation dataset. The test dataset is further divided into three test cases that help streamline the algorithm when tested across three different scenarios. Both the testing and validation dataset are subjected to be tested via the SVM algorithm. Max-pooling is performed in the dataset provided to the algorithm and through max-pooling. The computation cost can be reduced significantly on the one hand, while on the other hand, it helps in further isolating and extracting the highlighting features for prediction of onset of Type 2 Diabetes [5]. Then the LSTM architecture receives the max pooled feature set where the feature sets are passed onto the nodes of the input layer and the information received from the layer is combined in a weighted manner and the information is passed onto another input layer, and the input layer repeats the process until the information reaches the output layer [10]. The LSTM layer consists of 70 memory blocks that contain the capability to handle and dissect long-term dependencies in a data sequence. Then, the data is passed onto the RBF kernel for classification, which is used in conjunction with the SVM to provide the optimum outcome and maximum prediction accuracy for Type 2 Diabetes as well as slight improvement on the processing time for each subject in a dataset.

**Diabetes Prediction:** The features are then passed onto the RBF kernel coupled with the SVM algorithm to obtain maximum accuracy for diabetes prediction. The SVM and RBF kernel models are implemented using Scikit-learn, which performs a detailed 5-fold cross-validation on each of the feature set. Scikit-learn provides the outcome for each feature set as either diabetic or non-diabetic. The standard SVM has an exact separation on the dataset. However, it can cause the generalization of the dataset's features, which can cause the training data points to be misclassified and allowing to overlap class distribution [10]. Therefore, the RBF kernel is used to improve the generalization of the features, enabling the data to be transmitted through different input layers using a detailed 5-fold cross-validation. This helps to increase the accuracy of SVM and, at the same time, does not delay the processing time for the dataset. Equation 2 shows the RBF kernel's use with LSTM, which contains 70 memory blocks that learn the time domain features [10]. The outcomes from one sample are compared to another sample to identify the variable and differences amongst each other, improving the algorithm's accuracy and predicting the onset of Type 2 Diabetes better.

$$K(s, s1) = exp(\frac{||s - s_1||^2}{2\sigma^2}) \qquad (2)$$

Where,
$s_1$ = sample feature set 2
$s$ = sample feature set 1
$\sigma$ = RBF kernel variable

Using Eq. 2 from [10] enables to improve the accuracy of the state-of-the-art solution [1] by improving Eq. 1. It improves the accuracy of the SVM algorithm by replacing the slack variable ($\xi_i$) of the equation by LSTM layer, which is present inside the RBF kernel. The slack variable hampers the accuracy of the diabetes prediction and slows down the efficiency of prediction [10]. The RBF kernel provides better efficiency in predicting Type 2 Diabetes and improving the accuracy of prediction. Equation 3 shows the modified calculation expression.

$$MS = \min_{w, w_0, \xi} \frac{1}{2} ||w||^2 + C \sum_{i=1}^{M} MK(s, s1) \qquad (3)$$

Where,
$\xi_i$ = slack variable
$w_0$ = bias parameters
$w$ = LASSO penalty
$C$ = box constraint
$\sum_{i=1}^{M}(x_i, y_i)$ = training dataset of m observations and n features
$s_1$ = sample feature set 2
$s$ = sample feature set 1
$\sigma$ = RBF kernel variable

AUC is one of the important metrics in predicting Type 2 Diabetes, and the solution proposed by Nguyen et al. [8] helps improve the AUC metric for the prediction of onset of Type 2 Diabetes. It makes use of Multi-Layer Perceptron (MLP), which is a type of Artificial Neural Network. The feature sets are arranged in a multistage-directed graph to improve the AUC metric, as shown in Eq. 4.

$$y(x) = \sum_{i=1}^{M} w_i y_i(x) \qquad (4)$$

Where,
$y_i$ = output of network i
$w_i$ = associated weight

The final equation, Eq. 5, is obtained by combining Eq. 3 and Eq. 4, which combines the enhanced accuracy along with the improvement in the AUC metric, which helps to enhance the outcome of Eq. 1.

$$ES = MS + y(x) \qquad (5)$$

$$ES = \min_{w,w_0,\xi} \frac{1}{2}||w||^2 + C \sum_{i=1}^{M} \frac{||s - s_1||^2}{2\sigma^2} + \sum_{i=1}^{M} w_i y_i(x)$$

Where,
$\xi_i$ = slack variable
$w_0$ = bias parameters
$w$ = LASSO penalty
$C$ = box constraint
$\sum_{i=1}^{M}(x_i, y_i)$ = training dataset of m observations and n features
$s_1$ = sample feature set 2
$s$ = sample feature set 1
$\sigma$ = RBF kernel variable
$y_i$ = output of network i
$w_i$ = associated weight

The proposed solution provides an improvement of 7-9.5% on accuracy for predicting the onset of Type 2 Diabetes compared to the current industry standard. Likewise, the proposed solution also provides an improvement of 0.0341 or 3.41% on the AUC metric compared to the current industry standard and improves the processing time by 3.8 milliseconds compared to current industry standard.

### 4.1 Area of Improvement

Type 2 Diabetes is one of the most fatal and prominent diseases prevalent in today's modern society [11]. The use of machine learning and deep learning technique has drastically increased recently to predict the onset of Type 2 Diabetes. The proposed system uses a deep learning technique to predict the onset of Type 2 Diabetes to increase accuracy and improve AUC metrics. We have proposed Eq. 5, which improves the accuracy and AUC metric for predicting the onset of Type 2 Diabetes and improving the processing time compared to the current industry standard. The proposed equation improves Eq. 1 from the state-of-the-art solution [1] by combining the SVM algorithm with RBF kernel in the first stage to improve the accuracy of prediction of Type 2 Diabetes with the help of LTSM. With the SVM and RBF kernel help, max-pooling is performed on the features selected from datasets, which helps extract the sharpest and most relevant points from the datasets.

Similarly, LSTM is used to filter the input from the datasets, which helps obtain a better AUC metric [8]. The state-of-the-art system does not incorporate the RBF kernel. The proposed system incorporates the new kernel and LSTM layer and, at the same time, does not compromise the processing time and provides a better processing time in comparison to the current industry standard. As discussed in the literature review section, the available solutions that have been applied for the prediction of the onset of Type 2 Diabetes do not use the RBF kernel as well as it disregards the AUC metrics. Improving the accuracy and AUC metrics are vital to correctly and efficiently predict the onset of Type 2 Diabetes. With the struggle to find a perfect balance between the accuracy and AUC, the use of RBF kernel, LSTM, and MLP enable to obtain a better balance between these two important metrics, which can help in better prediction. Table 1 shows modified SVM with RBF kernel and LSTM layer.

Table 1 Modified Support Vector Machine with RBF kernel and LSTM layer

| Algorithm: Support Vector Machine (SVM) with Radial Base Function (RBF) Kernel and Long Short-term (LSTM) Layer |
|---|
| Input: feature sets obtained from the dataset, box constraint ($C$), associated weight ($w_i$) |
| Output: Prediction of onset of Type 2 Diabetes on any given subject |
| BEGIN<br>Step 1: Obtain the feature sets that have been filtered and selected to be analyzed and split the feature sets into testing dataset and training and validation dataset<br>Step 2: Binarize the values of feature sets so that the SVM algorithm can be initialized for data analysis<br>Step 3: Calculate the max-pooling value |

Step 4: Obtain the RBF kernel variable
Step 5: Take the feature sets through the LSTM layer for data validation, as shown in equation (4)
Step 6: Adjust the decision threshold in the validation feature set
Step 7: Decision Support System
Step 8: Display result
END

## 5. Results and Discussion

Python version 3.8.0 and the libraries like Tensorflow, Keras, and PyTorch frameworks were used to analyze and implement the proposed solution for the prediction of onset of Type 2 Diabetes [12]. Python was chosen as the tool of choice due to the open-source nature of the tool along with the large set of libraries available in Python and the usefulness of Python for soft computing, which makes it efficient and useful for the implementation of this research. Two different datasets are available for free to download from renowned sources, namely Global Diabetes Health Record, WHO and Pima Indians Dataset, and used to implement the proposed solution. The Pima Indians dataset was further divided into Pima Indians Dataset 1 and Pima Indians Dataset 2 to provide an accurate basis for implementing the research. The datasets were chosen to provide random feature sets that can enable the proposed implementation to be tested across different feature sets and provide variation in predicting the onset of Type 2 Diabetes. Whose features and characteristics have been presented below in Table 2 and 3. Adaptive Gradient Algorithm (AdaGrad) and a modified version of the stochastic gradient descent algorithm were used with a learning rate of 0.2 and a dropout rate of 0.35. The experimental system was powered by the machine running on AMD Ryzen™ 7 3700X CPU with an octa-core CPU clocked in at 3.4 GHz and 8.00 GB of LPDDR4 RAM.

Table 2. Statistics of the datasets chosen for the implementation of the proposed solution

| Dataset Name | Number of Samples | Number of Feature Sets | Selected Feature Sets |
|---|---|---|---|
| Pima Indians Dataset 1 | 769 | 9 | 8 |
| Pima Indians Dataset 2 | 768 | 9 | 8 |
| Global Diabetes Health Record, WHO | 2132 | 32 | 10 |

Table 3. Distribution of Positive and Negative Diabetic Patients in the dataset

| Dataset Name | Total Number of Samples | Positive Diabetic Patient | Negative Diabetic Patient |
|---|---|---|---|
| Pima Indians Dataset 1 | 769 | 562 | 207 |
| Pima Indians Dataset 2 | 768 | 497 | 271 |
| Global Diabetes Health Record, WHO | 2132 | 1365 | 767 |

The datasets were divided into two sub-datasets, namely, training dataset, testing and validation dataset. 30% of the dataset was used as a training dataset, and the remaining 70% was used as a testing and validation dataset. Comprehensive testing of each of the attributes of the dataset has been performed.

The input argument from Keras Model was used to create the first layer, which can be used to set to eight for the available. The eight variables were used to define the accuracy and AUC metrics and the system's processing time [13]. The architecture used Rectified Linear Unit Activation function (ReLU) for the first two layers and the sigmoid function for the output layer [5]. The sigmoid function was used in the outer layer to ensure the architecture's output is between 0 and 1, which in turn will facilitate easy prediction of onset of Type 2 Diabetes in any given subject with a default threshold that has been set to 0.5 [12].

The datasets used in the implementation of the new solution are converted to .csv format so that the proposed architecture can read and interpret the files without hiccups to calculate the accuracy and AUC metrics as well as the processing times for the prediction of onset of Type 2 Diabetes but without hampering the processing time [14]. The average accuracy and AUC metrics for each of the datasets and the state-of-the-art solution were calculated using evaluate() function, which calculates the mean value for the accuracy obtained for each dataset's prediction.

During the feature extraction and feature selection stage of data pre-processing, the datasets were subjected to feature extraction with the help of RBF, which works in conjunction with LSTM and then max-pooling is performed on the features selected from datasets. This helps to extract the sharpest and most relevant points from the datasets, which can help to increase the accuracy and AUC metric of Type 2 Diabetes prediction. The training dataset obtained from feature extraction is used to train the SVM Algorithm, and the testing and validation datasets are subjected to test for the prediction of onset of Type 2 Diabetes and the obtained data are subjected to further evaluation [8]. While calculating the updated accuracy and the AUC metrics, the processing times were also considered to ensure that the processing times do not take a hit in favour of prediction accuracy and the AUC metric.

The results obtained from the proposed solution as well as the state-of-the-art solution were compared and contrasted with the help of tables, bar graphs and line graphs. The results for both solutions extracted based on:

- The type of dataset (dataset with irregular data, small dataset and large dataset)
- The two different review types (positive and negative)

The outcome has been presented in two distinct categories: the dataset's division, namely, the training and testing dataset. The result for each category has been measured in terms of the prediction accuracy as well as the AUC metric and the processing time. The prediction accuracy refers to the total percentage of the correctly identified onset of Type 2 Diabetes amongst the test subject against the total number of available test subjects [15]. AUC metric refers to the rate of hitting the false alarm rate in any prediction model accuracy estimating the chosen positive instance, which is higher than a randomly chosen negative instance [16]. Epoch refers to the time taken to analyze one available test subject, where 1 epoch refers to the time elapsed to predict the onset of Type 2 Diabetes for that particular test subject [15]. The processing time refers to the amount of time taken by the proposed solution to predict the onset of Type 2 Diabetes in a given subject which is a part of the dataset. It is the total time elapsed between the time taken by the system first to interpret the data and provide an outcome for the prediction [17].

The positive and negative review for each of the datasets were tested separately for the state-of-the-art solution and the proposed solution. Ten samples were obtained from each of the selected datasets and the prediction accuracy, the AUC metric, and the processing time for each of the dataset was calculated for the state-of-the-art solution and the proposed solution. The test outcomes for both testing and training datasets have been presented in Table 4 to 10. The outcome of the test results has been presented in bar graphs in Figure 3 to 11.

The results were obtained by implementing the feature selection and classification stage of the proposed solution. The results were obtained by implementing the feature selection and classification stage of the proposed deep learning model. The proposed solution provides improved accuracy and AUC metric compared to the state-of-the-art solution with the help of Eq. 5, which combines of a deep learning technique that uses the SVM algorithm, the RBF, and the LSTM to predict the onset of Type 2 Diabetes. Likewise, the process uses an MLP to improve the AUC metric prediction. This helped in better and accurate prediction of the onset of Type 2 Diabetes in any given subject with the minimized risk of false positives and delayed outcomes, further proven by the improvement in the processing time, which lowers the processing times for the datasets. The industry standard diabetes prediction using machine learning techniques does not have a satisfactory prediction accuracy level [7]. Hence, the proposed system uses the RBF kernel coupled with the SVM algorithm to yield better prediction accuracy than the current industry standard machine learning approaches while keeping the processing time for each subject in check and improving the processing time.

Table 4. Accuracy and AUC metric for the state-of-the-art solution [1] and the proposed solution as per the chosen review datasets

| No. | Dataset | Stage | State-of-the-art solution | | | Proposed solution | | |
|---|---|---|---|---|---|---|---|---|
| | | | Accuracy (%) | AUC | Processing Time (epoch) | Accuracy (%) | AUC | Processing Time (epoch) |
| 1 | Global Diabetes Health Record, WHO | Training | 77.86% | 0.7652 | 14 | 85.62% | 0.8076 | 10 |
| | | Testing | 76.18% | 0.7980 | 12 | 88.18% | 0.8216 | 8 |
| 2 | Pima Indians Dataset 1 | Training | 77.81% | 0.7697 | 9 | 83.71% | 0.8123 | 5 |
| | | Testing | 78.80% | 0.7896 | 9 | 85.64% | 0.8319 | 7 |

| 3 | Pima Indians Dataset 2 | Training | 78.61% | 0.7765 | 11 | 86.21% | 0.8072 | 8 |
| | | Testing | 79.02% | 0.7912 | 9 | 87.81% | 0.8276 | 6 |

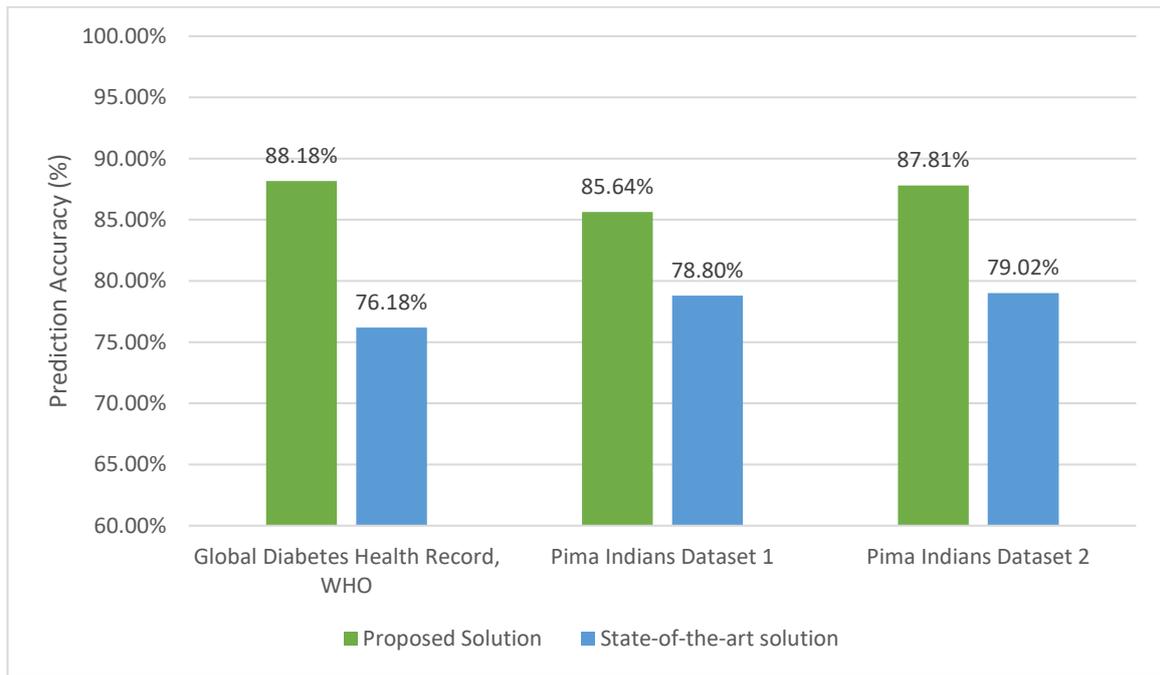

Figure 3. The prediction accuracy for three chosen datasets for the state-of-the-art solution [1] and the proposed solution during the testing stage.

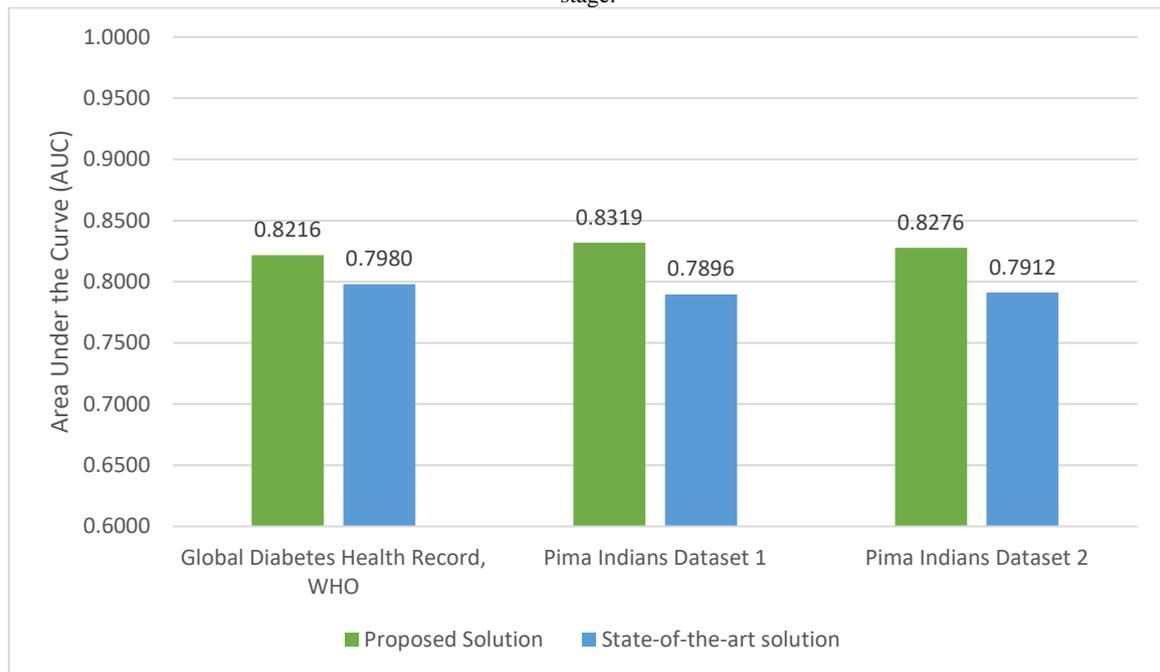

Figure 4. The AUC metric for three chosen datasets for the state-of-the-art solution [1] and the proposed solution during the testing stage.

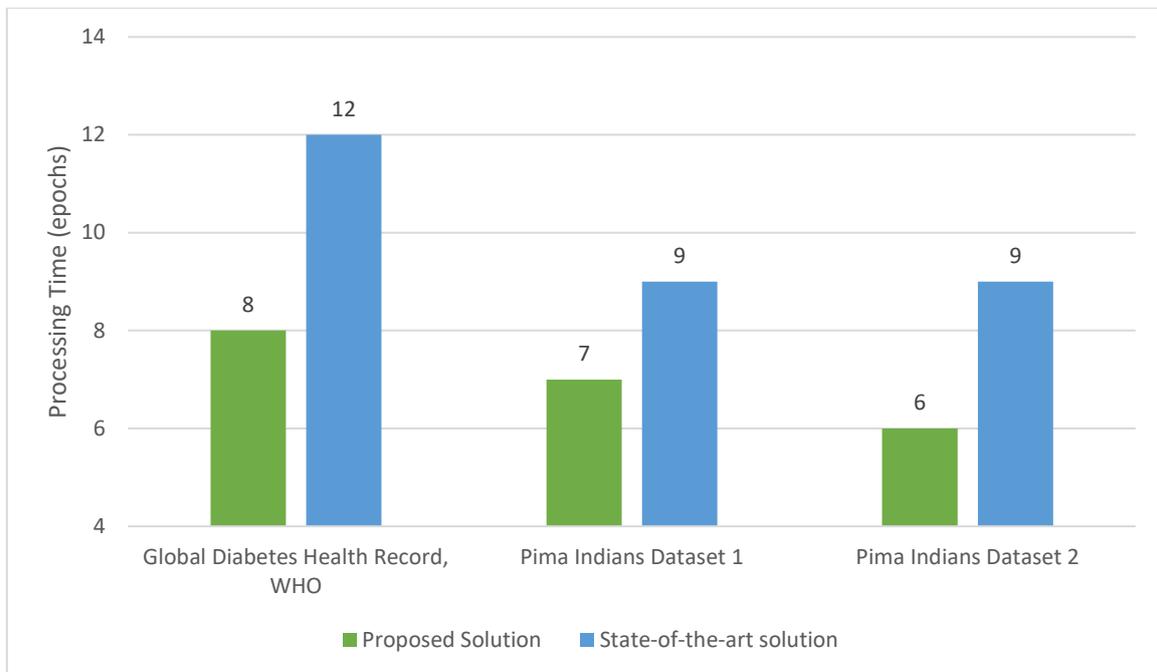

Figure 5. The processing time for three chosen datasets for the state-of-the-art solution [1] and the proposed solution during the testing stage.

Table 5. Accuracy and AUC Results for the State-of-the-art solution [1] and proposed solution for the positive review of Global Diabetes Health Record, WHO

| Sample No. | Sample Group | State-of-the-art solution | | Proposed solution | |
|---|---|---|---|---|---|
| | | Accuracy (%) | AUC | Accuracy (%) | AUC |
| 1.1 | Positive review from the Global Diabetes Health Record, WHO | 77.21 | 0.7862 | 88.97 | 0.8213 |
| 1.2 | | 76.24 | 0.7981 | 88.61 | 0.8451 |
| 1.3 | | 77.65 | 0.7789 | 87.62 | 0.8172 |
| 1.4 | | 78.91 | 0.7621 | 87.92 | 0.8218 |
| 1.5 | | 77.12 | 0.7768 | 88.31 | 0.8256 |
| 1.6 | | 78.15 | 0.7861 | 89.45 | 0.8311 |
| 1.7 | | 77.62 | 0.7767 | 89.29 | 0.8165 |
| 1.8 | | 77.17 | 0.7865 | 88.76 | 0.8101 |
| 1.9 | | 76.84 | 0.7766 | 88.44 | 0.8207 |
| 1.10 | | 77.87 | 0.7854 | 86.21 | 0.8217 |

Table 6. Accuracy and AUC Results for the State-of-the-art solution [1] and proposed solution for the negative review of Global Diabetes Health Record, WHO

| Sample No. | Sample Group | State-of-the-art solution | | Proposed solution | |
|---|---|---|---|---|---|
| | | Accuracy (%) | AUC | Accuracy (%) | AUC |
| 2.1 | Negative review from the Global Diabetes Health Record, WHO | 77.98 | 0.7972 | 87.12 | 0.8312 |
| 2.2 | | 76.87 | 0.7811 | 87.98 | 0.8412 |
| 2.3 | | 78.12 | 0.7923 | 88.21 | 0.8212 |
| 2.4 | | 77.71 | 0.7521 | 87.92 | 0.8329 |
| 2.5 | | 78.22 | 0.7718 | 87.39 | 0.8401 |
| 2.6 | | 78.87 | 0.7721 | 89.34 | 0.8232 |

| 2.7 | | 77.15 | 0.7707 | 89.98 | 0.8139 |
| 2.8 | | 78.12 | 0.7861 | 88.61 | 0.8101 |
| 2.9 | | 77.08 | 0.7781 | 88.50 | 0.8287 |
| 2.10 | | 77.87 | 0.7672 | 87.21 | 0.8209 |

Table 7. Accuracy and AUC Results for the State-of-the-art solution [1] and proposed solution for the positive review of Pima Indian Dataset 1

| Sample No. | Sample Group | State-of-the-art solution | | Proposed solution | |
|---|---|---|---|---|---|
| | | Accuracy (%) | AUC | Accuracy (%) | AUC |
| 3.1 | Positive review from the Pima Indians Dataset 1 | 77.12 | 0.7768 | 87.12 | 0.8312 |
| 3.2 | | 78.15 | 0.7861 | 88.61 | 0.8451 |
| 3.3 | | 76.62 | 0.7767 | 87.62 | 0.8172 |
| 3.4 | | 78.17 | 0.7865 | 87.92 | 0.8218 |
| 3.5 | | 76.84 | 0.7776 | 88.31 | 0.8256 |
| 3.6 | | 77.87 | 0.7834 | 89.45 | 0.8311 |
| 3.7 | | 76.85 | 0.7707 | 89.29 | 0.8165 |
| 3.8 | | 78.12 | 0.7861 | 88.61 | 0.8101 |
| 3.9 | | 77.08 | 0.7781 | 88.50 | 0.8287 |
| 3.10 | | 77.87 | 0.7672 | 87.21 | 0.8209 |

Table 8. Accuracy and AUC Results for the State-of-the-art solution [1] and proposed solution for the negative review of Pima Indians Dataset 1

| Sample No. | Sample Group | State-of-the-art solution | | Proposed solution | |
|---|---|---|---|---|---|
| | | Accuracy (%) | AUC | Accuracy (%) | AUC |
| 4.1 | Negative review from the Pima Indians Dataset 1 | 77.12 | 0.7768 | 88.50 | 0.8287 |
| 4.2 | | 78.15 | 0.7861 | 87.21 | 0.8209 |
| 4.3 | | 77.62 | 0.7705 | 87.62 | 0.8172 |
| 4.4 | | 77.17 | 0.7865 | 88.21 | 0.8291 |
| 4.5 | | 76.29 | 0.7709 | 87.92 | 0.8329 |
| 4.6 | | 78.22 | 0.7718 | 87.39 | 0.8401 |
| 4.7 | | 78.87 | 0.7721 | 89.29 | 0.8165 |
| 4.8 | | 77.98 | 0.7707 | 88.61 | 0.8101 |
| 4.9 | | 78.08 | 0.7781 | 88.31 | 0.8256 |
| 4.10 | | 79.87 | 0.7672 | 89.45 | 0.8311 |

Table 9. Accuracy and AUC Results for the State-of-the-art solution [1] and proposed solution for the positive review for Pima Indians Dataset 2

| Sample No. | Sample Group | State-of-the-art solution | | Proposed solution | |
|---|---|---|---|---|---|
| | | Accuracy (%) | AUC | Accuracy (%) | AUC |
| 5.1 | Positive review from the Pima Indians Dataset 2 | 77.17 | 0.7809 | 88.50 | 0.8212 |
| 5.2 | | 76.84 | 0.7732 | 87.92 | 0.8218 |
| 5.3 | | 79.87 | 0.7821 | 88.31 | 0.8221 |
| 5.4 | | 78.09 | 0.7783 | 88.21 | 0.8209 |
| 5.5 | | 76.26 | 0.7865 | 88.14 | 0.8312 |
| 5.6 | | 78.22 | 0.7718 | 88.32 | 0.8467 |
| 5.7 | | 78.87 | 0.7721 | 88.98 | 0.8421 |
| 5.8 | | 77.15 | 0.7707 | 88.09 | 0.8123 |
| 5.9 | | 78.62 | 0.7767 | 88.90 | 0.8432 |
| 5.10 | | 76.72 | 0.7865 | 89.76 | 0.821 |

Table 10. Accuracy and AUC Results for the State-of-the-art solution [1] and proposed solution for the negative review of Pima Indians Dataset 2

| Sample No. | Sample Group | State-of-the-art solution | | Proposed solution | |
|---|---|---|---|---|---|
| | | Accuracy (%) | AUC | Accuracy (%) | AUC |
| 6.1 | | 76.29 | 0.7612 | 87.92 | 0.8129 |
| 6.2 | | 78.22 | 0.7618 | 87.39 | 0.8501 |
| 6.3 | | 78.87 | 0.7821 | 89.29 | 0.8341 |
| 6.4 | | 77.09 | 0.7712 | 88.21 | 0.8321 |
| 6.5 | | 78.08 | 0.7891 | 88.31 | 0.8312 |
| 6.6 | Negative review from the Pima Indians Dataset 2 | 79.87 | 0.7782 | 89.45 | 0.8345 |
| 6.7 | | 78.87 | 0.7688 | 87.53 | 0.8421 |
| 6.8 | | 76.23 | 0.7612 | 86.12 | 0.8123 |
| 6.9 | | 78.02 | 0.7867 | 87.67 | 0.8232 |
| 6.10 | | 79.72 | 0.7865 | 89.09 | 0.821 |

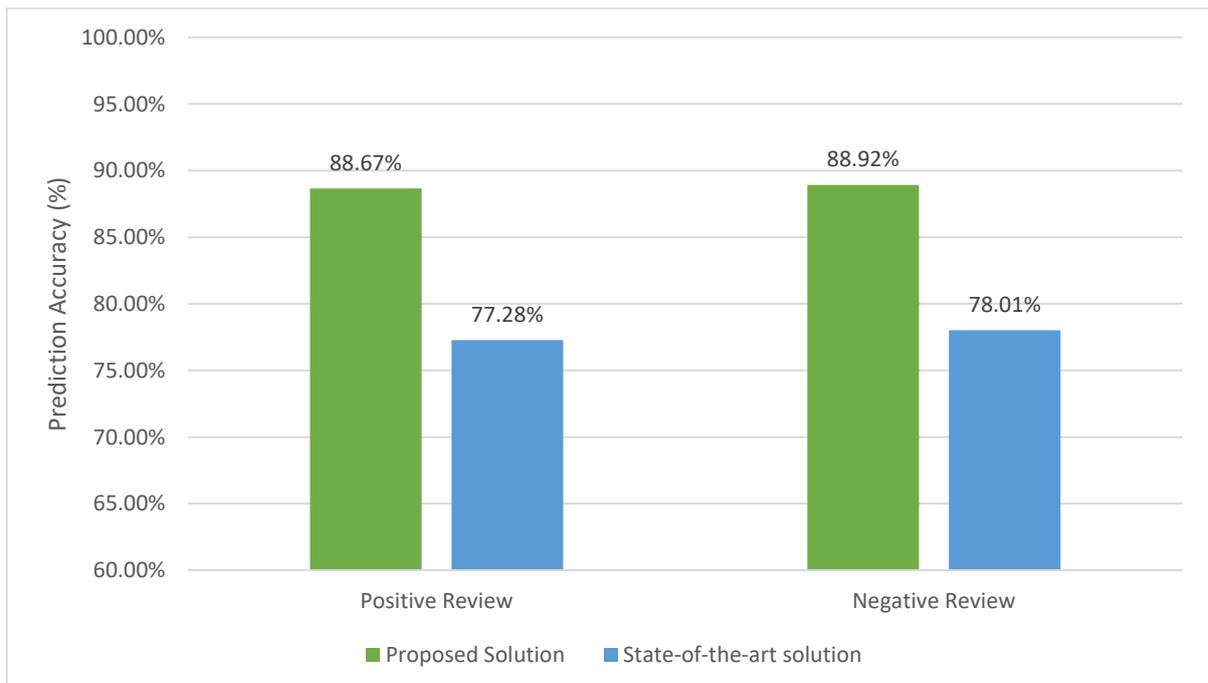

Figure 6. Average prediction accuracy in percentage for the positive and negative review for the state-of-the-art solution [1] and the proposed solution for Global Diabetes Health Record, WHO.

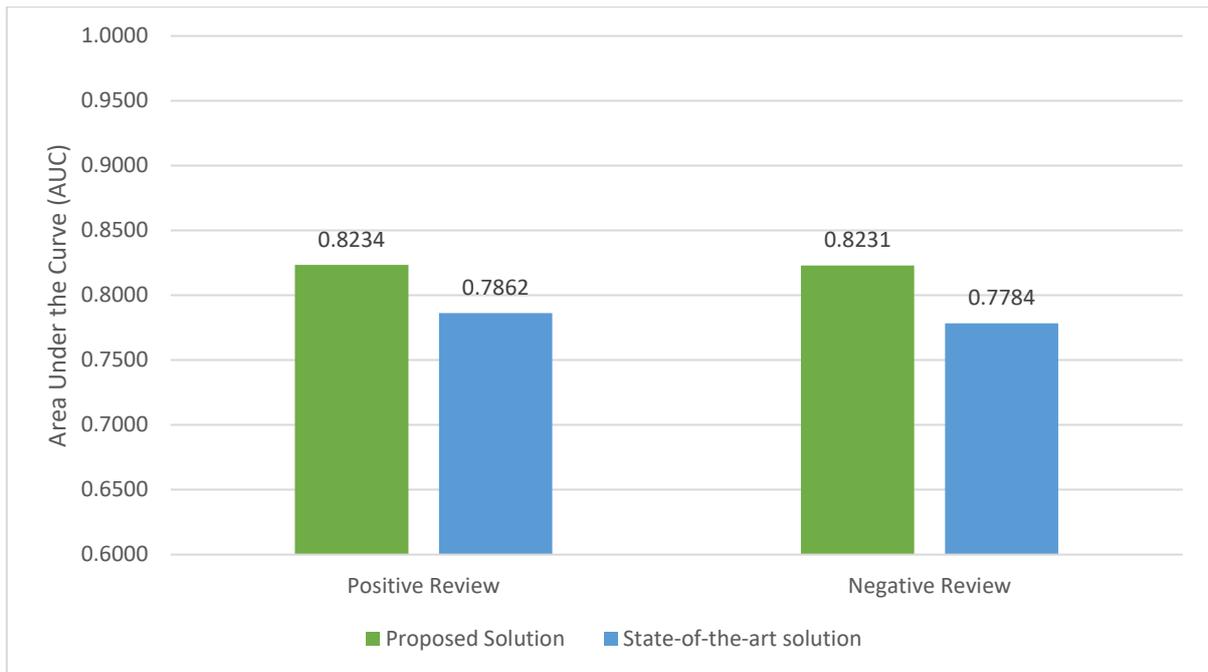

Figure 7. Average AUC for the positive and negative review for the state-of-the-art solution [1] and the proposed solution for Global Diabetes Health Record, WHO.

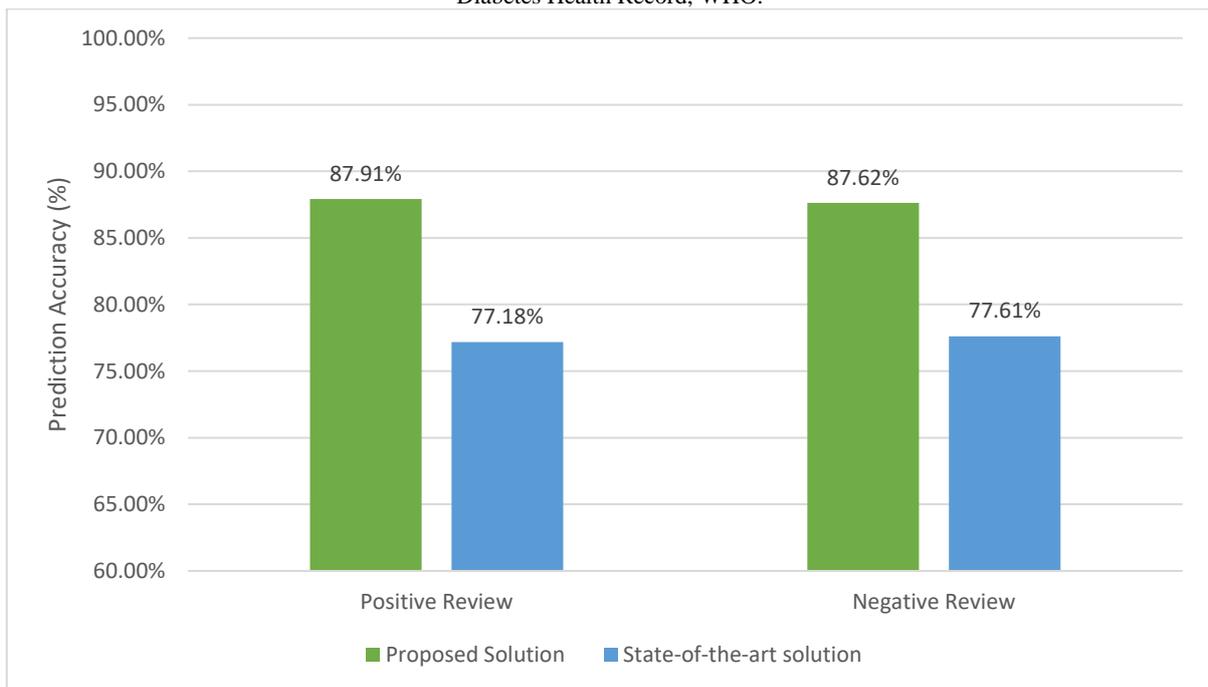

Figure 8. Average prediction accuracy in percentage for the positive and negative review for the state-of-the-art solution [1] and the proposed solution for Pima Indian Dataset 1.

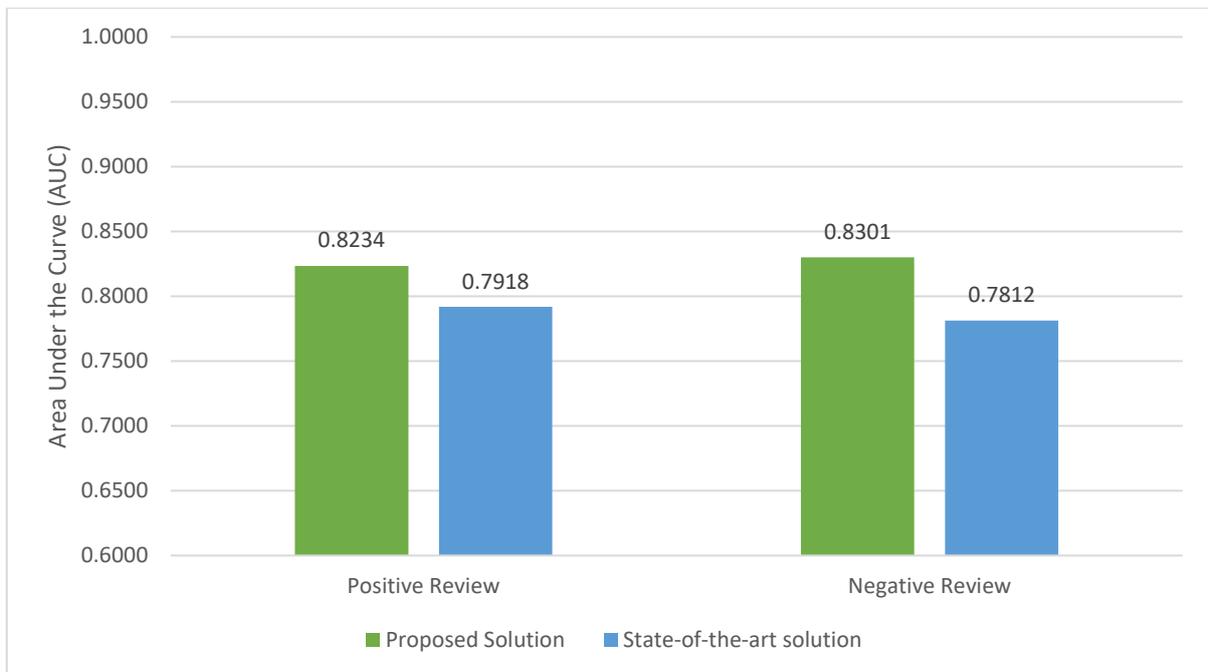

Figure 9. Average AUC for the positive and negative review for the state-of-the-art solution [1] and the proposed solution for Pima Indian Dataset 1

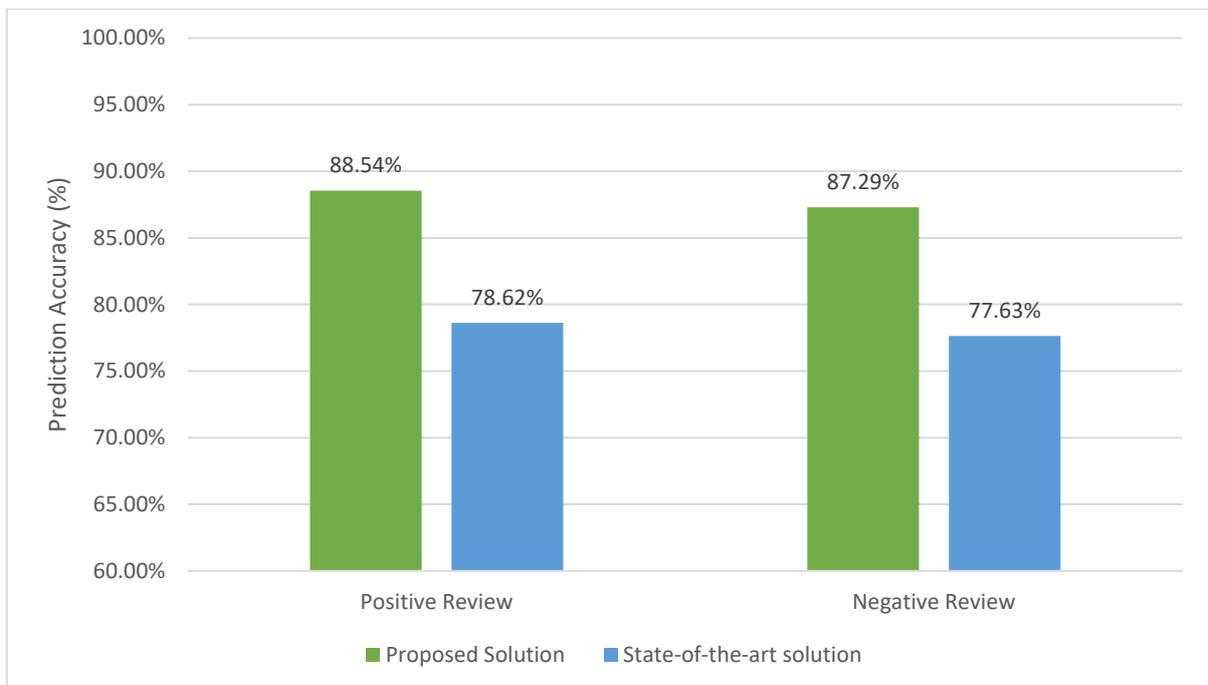

Figure 10. Average prediction accuracy in percentage for the positive and negative review for the state-of-the-art solution [1] and the proposed solution for Pima Indian Dataset 2.

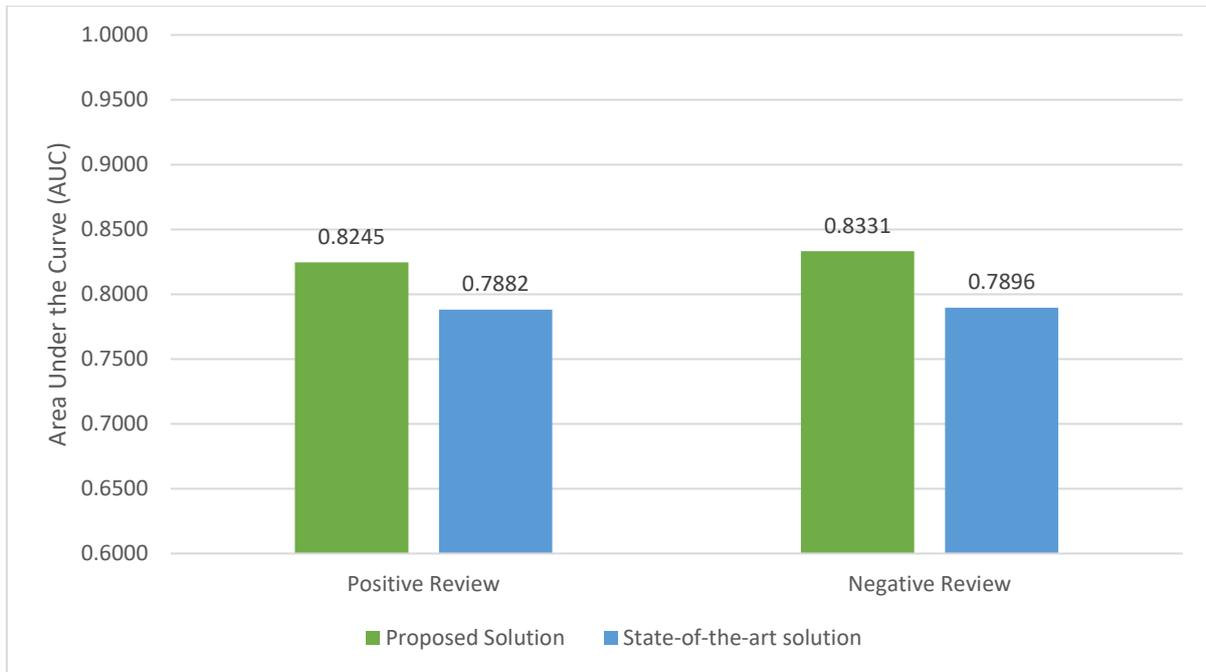

Figure 11. Average AUC for the positive and negative review for the state-of-the-art solution [1] and the proposed solution for Pima Indian Dataset 2.

The accuracy for each solution is obtained using the evaluate() function in Keras, a deep learning API within Python. It provides the true positive, true negative, false positive, and false negative values used to determine the accuracy and AUC metric for predicting the onset of Type 2 Diabetes. Once the datasets are analyzed individually, the average accuracy for the proposed solution and the state-of-the-art solution is obtained by using AVERAGE() function within Microsoft Excel. The improvement obtained in the proposed solution compared to the state-of-the-art solution was obtained by running both of the Algorithms with the Python framework tool's help and finding the difference between both solutions. The accuracy for predicting the onset of Type 2 Diabetes is calculated using Eq. 6 and AUC using Eq. 7.

$$Prediction\ Accuracy = \frac{(TP+TN)}{(TP+TN+FP+FN)} \qquad (6)$$

Where,
TP = True Positive
TN = True Negative
FP = False Positive
FN = False Negative

$$Area\ Under\ the\ ROC\ Curve\ (AUC) = \frac{TP}{TP+FP} \qquad (7)$$

Where,
TP = True Positive
FP = False Positive

The use of the RBF kernel in conjunction with the LSTM layer as proposed by Ryu et al. [10] and Nguyen et al. [8] improves prediction accuracy and the AUC value of the proposed solution. RBF kernel is introduced to the SVM to streamline the cross-validation accuracy of each of the subjects' feature set values, which helps to cut down the cross-validation errors and improve the overall accuracy of the algorithm. Likewise, the LSTM layer, as proposed by Ryu et al. [10], enables max-pooling on the dataset, which helps to improve the AUC metric of the system for further improvement on the accuracy. The use of the RBF kernel in conjunction with the SVM algorithm and the LSTM layer has helped cut down validation error and improve prediction accuracy. The accuracy is the most important metric while predicting the onset of Type 2 Diabetes because of the nature of the outcome that is necessary for the task, i.e. predicting the onset of a disease requires a high rate of accuracy [18] [19] [20]. The proposed solution has improved accuracy of 86.31%, which is a massive improvement over the current industry standard accuracy of 78.80%. Likewise, the AUC metric has also improved with an average AUC of 0.8270 or 82.70%, which is an improvement from the current industry standard AUC of 0.7929 or 79.29%. The processing time has also improved despite the addition of RBF kernel and LSTM layer with an average processing time of 42.2 milliseconds which has been reduced from the current industry standard of 45.9 milliseconds.

## 6. Conclusion and Future Work

Deep learning approaches have been on the rise for predicting the onset of Type 2 Diabetes, which can predict the onset with a satisfactory rate of accuracy and at a minimal cost and use of resources. This research focuses on the current industry standard deep learning approaches for predicting the onset of Type 2 Diabetes and providing further enhancement and improved accuracy while maintaining and improving the processing time. The modified SVM algorithm used in conjunction with the RBF kernel enables improvement in the accuracy and AUC metric for predicting the onset of Type 2 Diabetes using deep learning approaches compared to the state-of-the-art-solution. The proposed solution provides an average accuracy of 86.31% and an average AUC value of 0.8270 or 82.70%. RBF kernel and the LSTM layer enhance the prediction accuracy and AUC metric from the current industry standard, making it more feasible for practical use without compromising the processing time. The proposed solution improves the prediction accuracy by a massive 8.31% and the AUC metric by a respectable 3.41% concerning the current industry standard.

Similarly, the processing time improved to achieve 3.8 milliseconds in comparison to the current industry standard. This research used publicly available datasets and tested them in a small handpicked dataset, which puts the proposed solution's usability in an ideal real-life dataset into jeopardy. The results are predicted in a binary classification classified as either positive or negative (0 or 1). In the future, the proposed solution can be further enhanced and expanded to provide even better prediction accuracy and improve other metrics such as specificity and sensitivity of predicting the onset of Type 2 Diabetes.

**DECLERATION**
No Funding for this work and no Conflicts of interests as well

## References
[1] M. Bernardini, L. Romeo, P. Misericordia, and E. Frontoni, "Discovering the Type 2 Diabetes in Electronic Health Records Using the Sparse Balanced Support Vector Machine," *IEEE J Biomed Health Inform,* vol. 24, no. 1, pp. 235-246, Jan 2020, doi: 10.1109/JBHI.2019.2899218.
[2] S. Habibi, M. Ahmadi, and S. Alizadeh, "Type 2 Diabetes Mellitus Screening and Risk Factors Using Decision Tree: Results of Data Mining," *Glob J Health Sci,* vol. 7, no. 5, pp. 304-10, Mar 18 2015, doi: 10.5539/gjhs.v7n5p304.
[3] F. Mercaldo, V. Nardone, and A. Santone, "Diabetes Mellitus Affected Patients Classification and Diagnosis through Machine Learning Techniques," *Procedia Computer Science,* vol. 112, pp. 2519-2528, 2017, doi: 10.1016/j.procs.2017.08.193.
[4] Q. Zou, K. Qu, Y. Luo, D. Yin, Y. Ju, and H. Tang, "Predicting Diabetes Mellitus With Machine Learning Techniques," *Front Genet,* vol. 9, p. 515, 2018, doi: 10.3389/fgene.2018.00515.
[5] S. Islam Ayon and M. Milon Islam, "Diabetes Prediction: A Deep Learning Approach," *International Journal of Information Engineering and Electronic Business,* vol. 11, no. 2, pp. 21-27, 2019, doi: 10.5815/ijieeb.2019.02.03.
[6] T. Zheng *et al.*, "A machine learning-based framework to identify type 2 diabetes through electronic health records," *Int J Med Inform,* vol. 97, pp. 120-127, Jan 2017, doi: 10.1016/j.ijmedinf.2016.09.014.
[7] N. P. Tigga and S. Garg, "Prediction of Type 2 Diabetes using Machine Learning Classification Methods," *Procedia Computer Science,* vol. 167, pp. 706-716, 2020, doi: 10.1016/j.procs.2020.03.336.
[8] B. P. Nguyen *et al.*, "Predicting the onset of type 2 diabetes using wide and deep learning with electronic health records," *Comput Methods Programs Biomed,* vol. 182, p. 105055, Dec 2019, doi: 10.1016/j.cmpb.2019.105055.
[9] P. Samant and R. Agarwal, "Machine learning techniques for medical diagnosis of diabetes using iris images," *Comput Methods Programs Biomed,* vol. 157, pp. 121-128, Apr 2018, doi: 10.1016/j.cmpb.2018.01.004.
[10] K. S. Ryu, S. W. Lee, E. Batbaatar, J. W. Lee, K. S. Choi, and H. S. Cha, "A Deep Learning Model for Estimation of Patients with Undiagnosed Diabetes," *Applied Sciences,* vol. 10, no. 1, 2020, doi: 10.3390/app10010421.
[11] H. Wu, S. Yang, Z. Huang, J. He, and X. Wang, "Type 2 diabetes mellitus prediction model based on data mining," *Informatics in Medicine Unlocked,* vol. 10, pp. 100-107, 2018, doi: 10.1016/j.imu.2017.12.006.
[12] S. Raschka, J. Patterson, and C. Nolet, "Machine Learning in Python: Main Developments and Technology Trends in Data Science, Machine Learning, and Artificial Intelligence," *Information,* vol. 11, no. 4, 2020, doi: 10.3390/info11040193.
[13] R. Gast, D. Rose, C. Salomon, H. E. Moller, N. Weiskopf, and T. R. Knosche, "PyRates-A Python framework for rate-based neural simulations," *PLoS One,* vol. 14, no. 12, p. e0225900, 2019, doi: 10.1371/journal.pone.0225900.
[14] K. Kannadasan, D. R. Edla, and V. Kuppili, "Type 2 diabetes data classification using stacked autoencoders in deep neural networks," *Clinical Epidemiology and Global Health,* vol. 7, no. 4, pp. 530-535, 2019, doi: 10.1016/j.cegh.2018.12.004.
[15] Z. Xie, O. Nikolayeva, J. Luo, and D. Li, "Building Risk Prediction Models for Type 2 Diabetes Using Machine Learning Techniques," *Prev Chronic Dis,* vol. 16, p. E130, Sep 19 2019, doi: 10.5888/pcd16.190109.
[16] A. Cahn *et al.*, "Prediction of progression from pre-diabetes to diabetes: Development and validation of a machine learning model," *Diabetes Metab Res Rev,* vol. 36, no. 2, p. e3252, Feb 2020, doi: 10.1002/dmrr.3252.

[17]  T. R. Gadekallu *et al.*, "Early Detection of Diabetic Retinopathy Using PCA-Firefly Based Deep Learning Model," *Electronics,* vol. 9, no. 2, 2020, doi: 10.3390/electronics9020274.
[18]  A. Caliskan, Mehmet E. Yuksel, H. Badem, and A. Basturk, "Performance improvement of deep neural network classifiers by a simple training strategy," *Engineering Applications of Artificial Intelligence,* vol. 67, pp. 14-23, 2018, doi: 10.1016/j.engappai.2017.09.002.
[19]  A. Pimentel, A. V. Carreiro, R. T. Ribeiro, and H. Gamboa, "Screening diabetes mellitus 2 based on electronic health records using temporal features," *Health Informatics J,* vol. 24, no. 2, pp. 194-205, Jun 2018, doi: 10.1177/1460458216663023.
[20]  B. G. Choi, S. W. Rha, S. W. Kim, J. H. Kang, J. Y. Park, and Y. K. Noh, "Machine Learning for the Prediction of New-Onset Diabetes Mellitus during 5-Year Follow-up in Non-Diabetic Patients with Cardiovascular Risks," *Yonsei Med J,* vol. 60, no. 2, pp. 191-199, Feb 2019, doi: 10.3349/ymj.2019.60.2.191.
[21]  A. Mohebbi, T. B. Aradottir, A. R. Johansen, H. Bengtsson, M. Fraccaro, and M. Morup, "A deep learning approach to adherence detection for type 2 diabetics," *Conf Proc IEEE Eng Med Biol Soc,* vol. 2017, pp. 2896-2899, Jul 2017, doi: 10.1109/EMBC.2017.8037462.
[22]  B. J. Lee and J. Y. Kim, "Identification of Type 2 Diabetes Risk Factors Using Phenotypes Consisting of Anthropometry and Triglycerides based on Machine Learning," *IEEE J Biomed Health Inform,* vol. 20, no. 1, pp. 39-46, Jan 2016, doi: 10.1109/JBHI.2015.2396520.